\def\year{2021}\relax
\documentclass[letterpaper]{article} 
\usepackage{aaai21}  
\usepackage{times}  
\usepackage{helvet} 
\usepackage{courier}  
\usepackage[hyphens]{url}  
\usepackage{graphicx} 
\usepackage{natbib}  
\usepackage{caption} 
\usepackage{nohyperref} 
\usepackage[utf8]{inputenc} 
\usepackage[T1]{fontenc}    
\usepackage{url}            
\usepackage{booktabs}       
\usepackage{amsfonts}       
\usepackage{nicefrac}       
\usepackage{microtype} 
\usepackage{xcolor}
\usepackage{graphics}
\usepackage{mdwlist}
\usepackage{xspace}
\usepackage{amsmath}
\usepackage{mathtools}
\usepackage{wrapfig}
\usepackage[switch]{lineno} 
\usepackage{natbib} 
\usepackage{makecell}

\newcommand{\yset}{\mathrm{Y}}
\newcommand{\rset}{\mathrm{I}}

\newcommand{\featureFullName}{COVID-Augmented Exogenous Model\xspace}
\newcommand{\feature}{\textsc{Caem}\xspace}
\newcommand{\epicn}{\textsc{EpiDeep-CN}\xspace}
\newcommand{\epideep}{\textsc{EpiDeep}\xspace}
\newcommand{\empbayes}{\textsc{Empirical Bayes}\xspace}
\newcommand{\deltadensity}{\textsc{Delta Density}\xspace}

\newcommand{\gru}{\textsc{GRU}\xspace}
\newcommand{\prob}{COVID-ILI Forecasting Problem\xspace}
\newcommand{\ourmethod}{\textsc{Cali-Net}\xspace}
\newcommand{\supplement}{appendix}

\newtheorem{problem}{Problem}

    \makeatletter
\def\@fnsymbol#1{\ensuremath{\ifcase#1\or \mathsection\or \ddagger\or
   \mathsection\or \mathparagraph\or \|\or **\or \dagger\dagger
   \or \ddagger\ddagger \else\@ctrerr\fi}}
    \makeatother
\title{Steering a Historical Disease Forecasting Model Under a Pandemic:\\ Case of Flu and COVID-19}

\author{
    Alexander Rodr\'iguez\textsuperscript{\rm *}\thanks{equal contribution},
    Nikhil Muralidhar\textsuperscript{\rm $\dagger \mathsection$},
    Bijaya Adhikari\textsuperscript{\rm $\ddagger$},
    Anika Tabassum\textsuperscript{\rm *$\dagger$},\\
    Naren Ramakrishnan\textsuperscript{\rm $\dagger$},
    B. Aditya Prakash\textsuperscript{\rm *}\\
}
\affiliations{
    \textsuperscript{\rm *}College of Computing, Georgia Institute of Technology, Atlanta, USA\\
    \textsuperscript{\rm $\dagger$}Department of Computer Science, Virginia Tech, Arlington, USA\\
    \textsuperscript{\rm $\ddagger$}Department of Computer Science, University of Iowa, Iowa City, USA\\
    \{arodriguezc, badityap\}@gatech.edu, \{nik90, anikat1, naren\}@cs.vt.edu, bijaya-adhikari@uiowa.edu\\
}

\begin{document}
\maketitle

\begin{abstract}
Forecasting influenza in a timely manner aids health organizations and policymakers in adequate preparation and decision making. However, effective influenza forecasting still remains a challenge despite increasing research interest. It is even more challenging amidst the COVID pandemic, when the influenza-like illness (ILI) counts are affected by various factors such as symptomatic similarities with COVID-19 and shift in healthcare seeking patterns of the general population. 
Under the current pandemic, historical influenza models carry valuable expertise about the disease dynamics but face difficulties adapting. Therefore, we propose \ourmethod, a neural transfer learning architecture which allows us to 'steer' a historical disease forecasting model to new scenarios where flu and COVID co-exist. 
Our framework enables this adaptation by automatically learning when it should emphasize learning from COVID-related signals and when it should learn from the historical model. Thus, we exploit representations learned from historical ILI data as well as the limited COVID-related signals.
Our experiments demonstrate that our approach is successful in adapting a historical forecasting model to the current pandemic.
In addition, we show that success in our primary goal, adaptation, does not sacrifice overall performance as compared with state-of-the-art influenza forecasting approaches.
\end{abstract}

\section{Introduction}

Influenza is a seasonal virus which affects 9--45 million people annually in the United States alone resulting in between 12,000--61,000 deaths. Forecasting flu outbreak progression each year is an important and non-trivial task due to many confounding social, biological, and demographic factors. Accurate forecasts of the onset, peak and incidence can all aid significantly toward informing personalized policy roll out to minimize the effects of the flu season. To this end, the Centers for Disease Control and Prevention (CDC) has been organizing the FluSight challenge for the past several years, where the goal is to predict weighted influenza-like-illness counts (wILI) throughout the flu season in the United States~\cite{biggerstaff2016results}. wILI measures the percentage of healthcare seekers who show influenza like symptoms. Estimating various measures related to  the progression of a flu season (such as future incidence) gives policymakers valuable lead time to plan interventions and optimize supply chain decisions.

Moreover, the world has also been experiencing the devastating impacts of the COVID-19 pandemic which has sharply illustrated our enormous vulnerability to emerging infectious diseases. 
Hence in addition to being affected by various biological and demographic factors, wILI counts may now get further `contaminated' in the current (and possibly future) influenza seasons, in part due to symptomatic similarities with COVID.
\begin{figure*}[t]
  \centering
    \includegraphics[width = 0.8\linewidth]{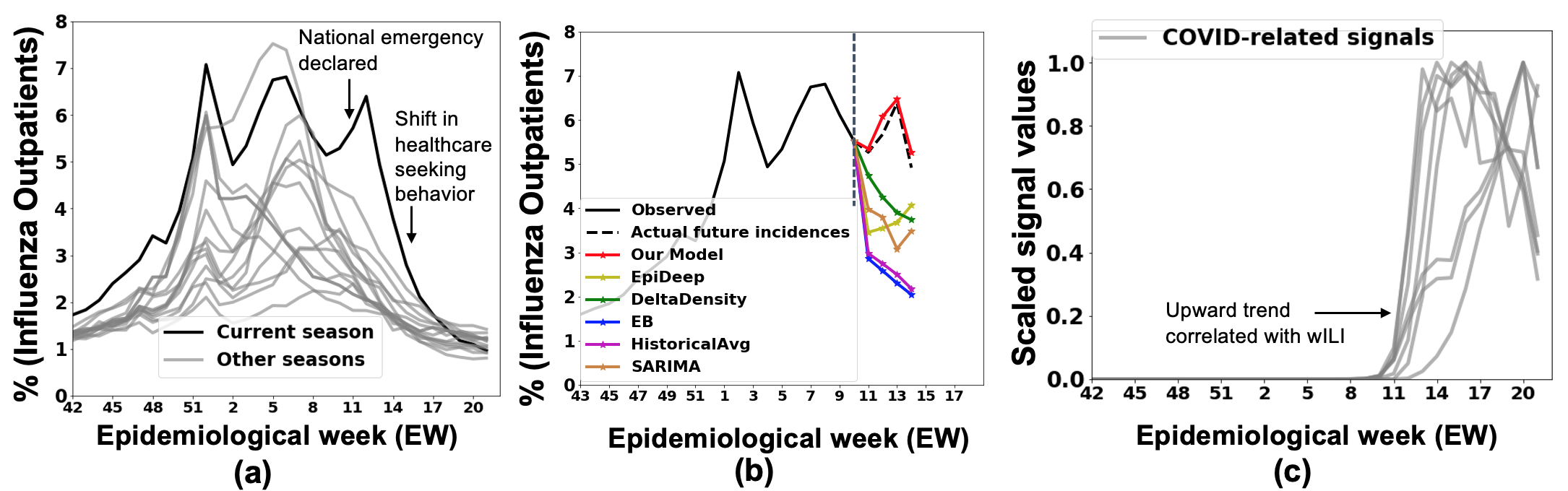}
  \caption{(a)  A novel forecasting scenario due to an emerging pandemic. Note the difference between the current and past seasons. (b) Established influenza forecasting methods are not able to adapt to uptrend caused by COVID. (c) Exogenous COVID related signals correlate better with wILI trend changes (due to contamination), which we exploit for more accurate forecasting.
  }
 \label{fig:intro}
\end{figure*}
Such miscounting manifests as significant changes in wILI seasonal progression as observed in Fig.~\ref{fig:intro}(a). Here the wILI curve for the current season 2019-2020 (contaminated by COVID, bold black) clearly shows a very different pattern compared to the previous seasons (in grey). To capture the deviance of the current wILI season and to forecast it in presence of COVID, we require a novel approach. Accurate forecasts of these unexpected trends in the current season are very helpful for resource allocation and healthcare worker deployment. Additionally, predicting wILI can also be used to help with indirect \textit{COVID surveillance}~\cite{castrofino2020influenza, boelle2020excess} - specially useful at the early stages of the pandemic, when there were no well established surveillance mechanisms for COVID. 
Finally, it is widely believed that COVID may be in circulation for a long period of time. Additionally, wILI itself models a mix of flu strains~\cite{cdc2020}. Hence, more generally, such a method can be used to disambiguate trends between historical strains and new emerging strains during a flu season.

There has been a recent spate of work on flu forecasting using statistical approaches usually trained on historical wILI data~\cite{adhikari2019epideep}. 
However, this new forecasting problem of adapting to a new emerging pandemic scenario is complex and cannot be addressed by traditional historical wILI methods alone. See Fig.~\ref{fig:intro}(b); current methods based on historical wILI cannot predict the uptrend, while our method (in red) can. 
The atypical nature of our `COVID-ILI' season may be caused by multiple co-occurring phenomena, e.g., the actual COVID-19 infections, the corresponding shutdowns and societal lock-downs and also changes in the healthcare-seeking behaviors of the public. This leads to the peak in COVID-ILI cases to be “out of step” with traditional wILI seasonal trends. As this feature is exclusive to this season, capturing  the new trend is a major challenge. 

Note that using only the historical wILI seasons is not sufficient to overcome it. Hence for this novel problem, we propose to leverage external COVID-related signals such as confirmed cases, hospitalizations, and emergency room visits as well. This leads us to the second challenge, viz. how to effectively model the COVID-ILI curve with new COVID-related signals, while also leveraging past prior knowledge present in the previous wILI seasons? However, note that these external signals are not available for historical wILI seasons. How do we address the imbalance in data to leverage both of these data sources? Further, as the contaminated COVID-ILI is a very new phenomenon which has suddenly emerged, there is limited data regarding the same from external signals and hence, a significant challenge is also to learn to model it effectively under data paucity.

To address these challenges, we propose \ourmethod (COVID Augmented ILI deep Network), a principled way to `steer' flu-forecasting models to adapt to new scenarios where flu and COVID co-exist. We employ transfer learning and knowledge distillation approaches to ensure effective transfer of knowledge of the historical wILI trends. We incorporate multiple COVID-related data signals all of which help capture the complex data contamination process showcased by COVID-ILI. As shown in Fig.~\ref{fig:intro}(c), these exogenous signals correlate better with the anomalous trends caused by COVID.  Finally, in order to alleviate the data paucity issue, we train a single global architecture with explicit spatial constraints to model COVID-ILI trends of all regions as opposed to previous approaches which have modeled each region separately leading to a superior forecasting performance (See Sec.~\ref{sec:experiments}).

Our contributions are as follows:

\begin{enumerate*}
    \item We develop~\ourmethod, a novel heterogeneous transfer learning framework to adapt a flu forecasting historical model into the new scenario of COVID-ILI forecasting.
    \item We embed~\ourmethod with a recurrent neural network including domain-informed spatial constraints to capture the spatiotemporal dynamics across different wILI regions.
    \item We also employ a Knowledge Distillation scheme to explicitly transfer historical wILI knowledge to our target model in~\ourmethod, thereby alleviating the effect of paucity of COVID-ILI data.
    \item Finally, we show how \ourmethod succeeds in adaptation, and also
    perform a rigorous performance comparison of~\ourmethod with several state-of-the-art wILI forecasting baselines. In addition, we perform several quantitative and qualitative experiments to understand the effects of various components of~\ourmethod.
\end{enumerate*}
Overall, more broadly, our work is geared towards adapting a historical model to an emerging disease scenario, and we specifically demonstrate the effectiveness of our approach in the context of wILI forecasting in the COVID-19 emerging disease scenario. Appendix, code, and other resources can be found online\footnote{Resources website: \url{https://github.com/AdityaLab/CALI-Net}}\!.

\section{Related Work}
To summarize, we are the first to address the problem of adapting to shifting trends using transfer learning and knowledge distillation in an epidemic forecasting setting, leveraging exogenous signals as well as historical models. Our research draws from multiple lines of work.  

\par \noindent \textbf{Epidemic Forecasting: } Several approaches for epidemic forecasting have been proposed including statistical \cite{tizzoni2012real,adhikari2019epideep,osthus2019dynamic,brooks_nonmechanistic_2018}, mechanistic \cite{shaman2012forecasting,zhang2017forecasting}, and ensemble \cite{reich2019accuracy} approaches. Several approaches rely on external signals such as environmental conditions and weather reports \cite{shaman2010absolute,tamerius2013environmental,volkova2017forecasting}, social media \cite{chen2016syndromic,lee2013real}, search engine data \cite{ginsberg2009detecting,yuan2013monitoring}, and a combination of multiple sources \cite{chakraborty2014forecasting}.
Recently, there has been increasing interest in deep learning for epidemic forecasting \cite{adhikari2019epideep,wang2019defsi,rodriguez2020deepcovid}. These methods typically exploit intra and inter seasonal trends. Other approaches like \cite{venna2017novel} are limited to specific situations, e.g., for military populations. However, to the best of our knowledge, there has been no work on developing deep architectures for adapting to trend shifts using exogenous data.

\par \noindent \textbf{Time Series Analysis: } 
There are several data driven, statistical and model-based approaches that have been developed for time series forecasting such as auto-regression, Kalman-filters and groups/panels \cite{box2015time,sapankevych2009time}. 
Recently, deep recurrent architectures \cite{hochreiter1997long} have shown great promise in learning good representations of temporal evolution \cite{fu2016using,muralidhar2019dyat,connor1994recurrent}. 

\par \noindent \textbf{Transfer Learning within heterogeneous domains: } 
This challenging setting of transfer learning with heterogeneous domains (different feature spaces)  
aims to leverage knowledge extracted from a \emph{source} domain to a different but related \emph{target} domain.
\cite{moon_completely_2017} proposed to learn feature mappings in a common-subspace, and then apply shared neural layers where the transfer would occur. 
\cite{li_heterogeneous_2019} proposed transfer learning via deep matrix completion.
\cite{yan_semi-supervised_2018} formulated this problem as an optimal transport problem using the entropic Gromov-Wasserstein discrepancy. 
We adapt the classification method in~\cite{moon_completely_2017} to our regression setting, for effectively transferring knowledge from the source to the target model in our COVID-ILI forecasting task.
Knowledge Distillation (KD) is also a popular transfer learning method, to develop shallow neural networks capable of yielding performance similar to deeper models by learning to "mimic" their behavior~\cite{ba2014deep,hinton2015distilling}.  \cite{saputra2019distilling} inspects KD for deep pose regression. The authors propose two regression specific losses, namely the hint loss and the imitation loss which we adapt in this work for  COVID-ILI forecasting. Unlike our paper, most KD work has been applied to classification and efforts for adapting KD for regression have been sparse~\cite{saputra2019distilling,takamoto2020efficient}.

\section{Background}

\noindent \textbf{COVID-ILI forecasting task:} 
Here we consider a short term forecasting task of predicting the next \textit{k} wILI incidence given the data till week $t-1$ for each US HHS region and the national region. This corresponds to predicting the wILI values for week $\{t$, $t+1$, \ldots, $t+k\}$ at week $t$ (matching the exact real-time setting of the CDC tasks) for each region.

We are given a set of historical annual wILI time-series,  $\yset_i  = \{y^1_i, y^2_i, \dots, y^{t-1}_i \}$ for each region $i$. The wILI values have been contaminated by COVID-19 for all weeks $t \geq w$. We also have various COVID-related exogenous data signals $X_i = \{ \mathbf{x}_i^{w}, \mathbf{x}_i^{w+1}, \dots, \mathbf{x}_i^{t-1} \}$, where each feature vector $\mathbf{x}^j_{i}$ is constructed using various signals such as COVID line list data, test availability, crowd-sourced symptomatic data, and social media. 
Our task is to forecast the next \textit{k} wILI incidence for all regions $i \in \rset$. Specifically, our novel problem is:

\begin{problem} \prob \\
\textbf{Given}: a set of historical annual wILI time-series $\yset_i =  \{y_i^1, y_i^2, \dots, y_i^{t-1} \}$ for regions $i \in \rset$ and the set of COVID-related exogenous signals  $X_i = \{ \mathbf{x}_i^{w}, \mathbf{x}_i^{w+1}, \dots, \mathbf{x}_i^{t-1} \}$ for the current season and regions $i \in \rset$. \\
\textbf{Predict:} next k wILI incidences $\forall_{j = t}^{t+k} y_i^j$ for each region $i \in \rset$.
\end{problem}

\par \noindent \textbf{Epideep for wILI forecasting. }
\epideep \cite{adhikari2019epideep} is a deep neural architecture designed specifically for wILI forecasting. The core idea behind \epideep is to leverage the seasonal similarity between the current season and historical seasons to forecast various metrics of interest such as next incidence values, onset of current season, the seasonal peak value and peak time. To infer the seasonal similarity between current and the historical season, it employs a deep clustering module which learns latent low dimensional embeddings of the seasons. 

\section{Our Approach}

In this section, we describe our method \ourmethod, which models COVID-ILI by incorporating historical wILI knowledge as well as the limited new COVID-related exogenous signals. 
Next we give an overview of how our approach  uses heterogeneous transfer learning (HTL), overcomes data paucity issues, and controls the transfer of only useful knowledge and avoids negative transfer.

\subsection{Exploiting Learned Representations from Historical wILI via HTL}
\label{subsection:HTL}

We leverage recent advances in HTL to incorporate the rich historical wILI data. To that end, we use the \epideep model as our base model. \epideep{} was designed to learn representations from historical wILI that embed seasonal and temporal patterns. Here, we adapt the CTHL framework~\cite{moon_completely_2017} to transfer knowledge from \epideep. 

\begin{figure*}[t]
    \centering
    \includegraphics[width = 0.73\linewidth]{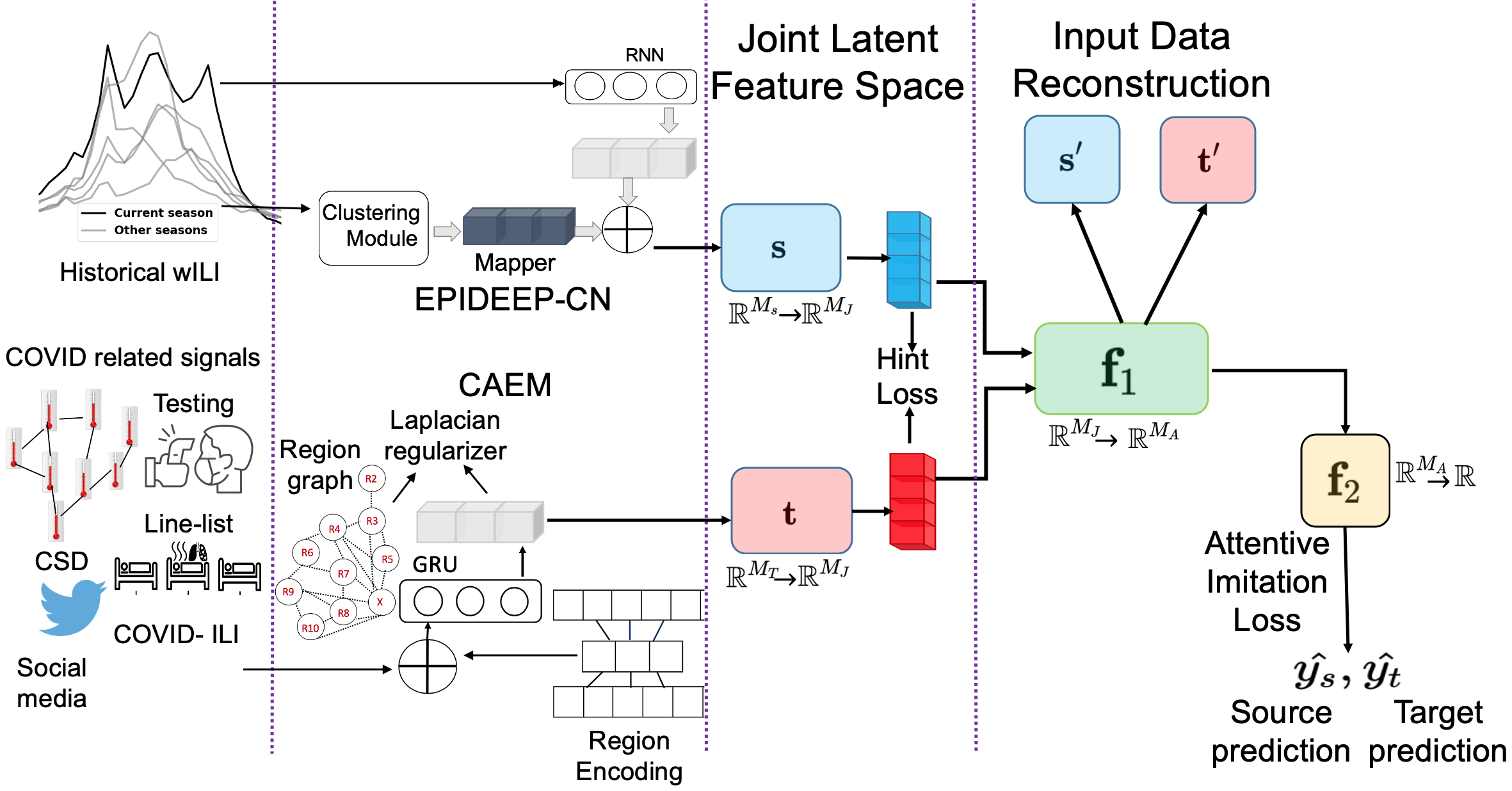}
    \caption{Our proposed model \ourmethod. 
    Our heterogeneous transfer learning architecture is designed to transfer knowledge from \epicn about historical wILI trends to the \feature module (using exogenous signals) for COVID-ILI forecasting while addressing the challenges of negative transfer, spatial consistency and data paucity.} 
    \label{fig:model}
\end{figure*}

In our HTL setting, a modified version of the \epideep model is the source model and we design a feature module (discussed in Sec.~\ref{subsec:regional}) to be the target model.
As depicted in Fig.~\ref{fig:model}, the embeddings of the source and target are each transformed by modules $\textbf{s}$ and $\textbf{t}$ respectively, such that the latent embeddings of the source and target model are placed into a common feature space. In this way, we are projecting knowledge extracted from both heterogeneous feature spaces into a shared latent space. Formally, the transformations may be expressed as $\mathbf{s}: \mathbb{R}^{M_{S}} \rightarrow \mathbb{R}^{M_{J}}$ and $\mathbf{t}: \mathbb{R}^{M_{T}} \rightarrow \mathbb{R}^{M_{J}}$, where $M_S$ and $M_T$ are the dimensions of the source and target embeddings, respectively, and $M_{J}$ is the dimension of the joint latent feature space. After projecting representations from the source and the target models into a joint latent feature space, a sequence of shared transformations $\textbf{f}_1:\mathbb{R}^{M_{J}} \rightarrow \mathbb{R}^{M_{A}}$ and $\textbf{f}_2:\mathbb{R}^{M_{J}} \rightarrow \mathbb{R}$ is applied on them, thereby transporting them both into the same latent space. 
On top of this architecture, we employ a denoising autoencoder to reconstruct the input data as we find it improves our latent representations.  
These modules are depicted as $\textbf{s}'$ and $\textbf{t}'$. 

\subsection{COVID-Augmented Exogenous Model (\feature)}
\label{subsec:regional}
Our target model from Sec. \ref{subsection:HTL} could be a simple feedforward network. Instead, 
to alleviate the data paucity that exists for the COVID-related exogenous data, we develop the~\featureFullName{} (\feature) which \emph{jointly} models all regions exploiting regional interplay characteristics.
Such an approach allows us to extract the most out of our limited training data enabling us to employ more sophisticated sequential architectures.
To enable model awareness of multiple regional patterns, we explicitly encode each \textit{region embedding} $\mathbf{r}\in \mathbb{R}^{1\times h_r}$, and pass to~\feature{} along with the exogenous input data of the region for a particular week. The region embeddings are produced by an autoencoder whose task is to reconstruct one-hot encodings of each region.

The data we consider exhibit sequential dependencies. In order to model these dependencies, we employ the popular GRU recurrent neural architecture ~\cite{cho2014learning}. The GRU is trained to encode temporal dependencies using data from week $t-W$ to week $t-1$ and predict values for week $t+k$. 
At each step of recurrence, the GRU receives as input, exogenous data signals $\mathbf{x}^{t-\lambda}_i \in \mathbb{R}^{1\times l}$ (for week $t-\lambda$ with $\lambda \in \{W,W-1,\ldots,1\}$ and region $i$) and the region embedding $r_i$, both concatenated to form the full GRU input. For simplicity, henceforth, we consider $\mathbf{x}^{t-\lambda}_i \in \mathbb{R}^{1\times l+h_r}$ to represent this concatenated input to the GRU ($l$ is the number of different data signals we employ and $h_r$ is the dimension of the latent region embedding obtained from the \feature{} Region Embedding autoencoder).

\par\noindent
\textbf{Laplacian Regularization:}
Infectious diseases like COVID and flu naturally also show strong spatial correlations and to capture this aspect of the wILI season evolution across different regions effectively, we incorporate spatial constraints using Laplacian Regularization~\cite{belkin2004regularization} and predict COVID-ILI values for all regions jointly. Let us consider the region graph $G(V,E)$ where $V$ (vertices) indicates the number of regions (11 in our case including the national region) and $E$ indicates edges between the vertices. Two regions are considered to have an edge between them if they are bordering each other. We construct $G$ based on region demarcations provided by the HHS/CDC and connect the national region to all other regions.

The optimization objective for \feature is as follows:
\begin{equation}
\begin{aligned}
    \mathrm{min}_{\Theta_{RE}\Theta_{F}}\, &||F(\mathbf{X}^{t-W:t-1};\Theta_F) - Y^{t}||_2^2 +\\ \,&RE(\textbf{E};\Theta_{RE}) + Tr(\textbf{h}^T\textbf{L}\textbf{h})
    \label{eq:feature_module_optimization}
\end{aligned}
\end{equation}

In Eq.~\ref{eq:feature_module_optimization}, $\Theta_{RE},\Theta_F$ represent the model parameters for \textit{region embedding} (RE) function and the recurrent \textit{forecasting} (F) function respectively. The input to F, $\textbf{X}^{t-W:t-1} \in R^{|V|\times l+h_r}$, is the historical COVID-related exogenous data for the past $W$ weeks for all 11 regions (along with the regional embedding for each). The output of F, $Y^t \in R^{|V|\times 1}$ for week $t$ includes forecasts for all 11 regions. The region embedding is generated using an autoencoder which accepts the one-hot encoding for all regions $\textbf{E}\in \mathbb{B}^{|V|\times|V|}$. Finally, $\mathbf{h} \in \mathbb{R}^{1\times h_r}$ is the hidden representation of the input sequence generated by the forecasting model F at the end of the recurrence which is used to enforce regional representation similarity governed by the normalized Laplacian (\textbf{L}) of graph G. Laplacian regularization has been shown to systematically enforce regional similarity, effectively capturing spatial correlations~\cite{subbian2013climate}. Both the RE and F modules of \feature are jointly trained coupled with Laplacian regularization. It must be noted that when integrated into \ourmethod, the function $F$ includes the GRU parameters and the parameters for transformations $\textbf{t}$,$\,\textbf{f}_1$,$\,\textbf{f}_2$ employed to yield the final k-week ahead predictions.

\subsection{Attentive Knowledge Distillation Loss}
\label{subsec:KD}

A mechanism for the target model to exercise control over knowledge transfer and prevent negative transfer is necessary in our setting to avoid the transfer of possibly erroneous predictions made by \epideep for the atypical portions of the current influenza season. 
To enable this, we employ attentive knowledge distillation (KD) techniques. Recently,~\cite{saputra2019distilling} has employed KD in deep pose estimation. We noticed that our modification to this method is capable of not only transferring knowledge from \epideep (our source model) to \feature (our target model) but also showcases how to selectively transfer knowledge based on the quality of source model predictions. 

\noindent\textbf{Adapting \epideep{}:} In order to achieve effective transfer of knowledge from \epideep to \feature, we modify the existing \epideep architecture.
\epideep, by design, requires a different model to be trained for each week in the season which is not ideal for effective knowledge transfer. Hence, to prevent this, we modify \epideep into \epicn (EpiDeep-\ourmethod) which incrementally re-trains the same model for each week in the season, thereby allowing the KD losses to be applied to the same set of \epicn parameters, enabling more efficient knowledge transfer to \feature. Specifics about \epicn are in our \supplement. 

The training datasets for \epicn and \feature are not the same, however they share an overlapping subset of data, from January 2020 onward when the COVID pandemic started. Therefore, we enforce the \emph{KD loss} only on this subset of the training data.
Our KD loss is composed of two terms: imitation loss $\mathcal{L}_{\text{Im}}$ and hint loss $\mathcal{L}_{\text{Hint}}$, described mathematically as follows,
\begin{equation}
\begin{aligned}
\mathcal{L}_{KD} &= 
\alpha \frac{1}{n} \sum_{i=1}^{n}
\Phi_{i} \underbrace{ \left\| \hat{y}_{s} -\hat{y}_{t}\right\|_{i}^{2}}_\text{$\mathcal{L}_{\text{Im}}$} ~+~
\Phi_{i}
\underbrace{\left\|
\Psi_{s} - \Psi_{t} \right\|_{i}^{2}}_\text{$\mathcal{L}_{\text{Hint}}$} \\
\end{aligned}
\end{equation}
where $\Phi_{i} =\left(1-\frac{\left\|\hat{y}_{s}-y\right\|_{i}^{2}}{\eta} \right)$, and   $\Psi_{s}$ and $\Psi_{t}$ are the output embeddings of $\textbf{s}$ and $\textbf{t}$, respectively;
$i\in \{1,\ldots,n\}$ is the index for each training observation, and $n$ the batch size;
${\eta =\max \left(e_{s}\right)-\min \left(e_{s}\right)}$ is a normalizing factor (i.e range of squared error losses of source model) and ${e_{s} =\left\{\left\|y - \hat{y}_{s}\right\|_{j}^{2}: j=1, \ldots, N\right\}}$, the actual set of squared errors between the source predictions and ground truth.
$N$ is the total number of observations in the overlap training data; $\Phi_{i}$ is the attention weight assigned to the $i^{th}$ training observation. The attention is a function of how well the source model is able to predict ($\hat{y}_s$) a particular ground truth target $y$. The attention weights are applied over the imitation loss between the source predictions ($\hat{y}_s$) and the ground truth ($\hat{y}_t$) as well as over the hint loss between the latent output embeddings $\Psi_s$ and $\Psi_t$ to ensure transfer of knowledge at multiple levels in the architecture from the source \epicn to the target \feature. The goal of KD is to enforce a unidirectional transfer from the source (\epicn) to the target (\feature) model. Hence KD losses do not affect the representations learned by \epicn and module \textbf{s}.

\section{Experiments}
\label{sec:experiments}

\begin{table*}[t]
    \centering
    \caption{Overview of COVID-Related Exogenous Data.}
    {    \small
    \begin{tabular}{|l|l|l|l|}
    \hline
    {\bf Type of signal} & \textbf{Description} & {\bf Signals} & {\bf Source}\\\hline
    
    (DS1) Line list & They are a  &  1. Confirmed cases; 2. UCI beds;& \cite{covidtracking,cdc2020} \\
     based & direct function & 3. Hospitalizations; 4. People on& \cite{jhu-covid} \\
      & of the disease spread & ventilation; 5. Recovered; 6. Deaths;&  \\
      &     & 7. Hospitalization rate; & \\
      &     & 8. ILI ER visits; 9. CLI ER visits & \\\hline
    (DS2) Testing & Related to social & 10. People tested; 11. Negative cases; & \cite{covidtracking,cdc2020}\\
    based &  policy and behavioral & 12. Emergency facilities reporting;& \\
          & considerations & 13. No. of providers; &  \\\hline
(DS3) Crowdsourced & Crowdsourced symptomatic  & 14. Digital thermometer readings;&  \cite{miller2018smartphone}\\
symptoms based & data from personal devices & & \\\hline
(DS4) Social media  & Social media activity & 15. Health Related Tweets & \cite{dredze2014healthtweets} \\\hline
    \end{tabular}
    \label{tab:datasets}
    }
\end{table*}
\par\noindent

\noindent\textbf{Setup}. All experiments are conducted using a 4 Xeon E7-4850 CPU with 512GB of 1066 Mhz main memory. Our method implemented in PyTorch (implementation details in the appendix) is very fast, training for one predictive task in about 3 mins. Here, we present our results for next incidence prediction (i.e. $k = 1$). We present results for next-two incidence predictions in the \supplement, which are similar. Note that we define $T_1$ as the period of non-seasonal rise of wILI due to contamination by COVID-19 related issues (EWs 9-11), $T_2$ as the time period when COVID-ILI trend is declining more in tune with the wILI pattern (EWs 12-15), and $T$ as the entire course (EWs 9-15).

\noindent\textbf{Data.} We use the historical weighted Influenza-like Illness (wILI) data released by the CDC which collects it through the Outpatient Influenza-like Illness Surveillance Network (ILINet). ILINet consists of more than 3,500 outpatient healthcare providers all over the US. We refer to wILI from June 2004 until Dec 2019 as \textbf{historical wILI}, and wILI from January 2020 as \textbf{COVID-wILI}. Next, Table~\ref{tab:datasets} details the various signal types we employ for COVID-related exogenous data. A more detailed description of each data signal can be found in the \supplement. All datasets are publicly available and were collected in May 2020.

\noindent\textbf{Goals.} In our experiments we aim to demonstrate that our method \ourmethod can systematically steer a historical model to the new COVID-ILI scenario by enabling it to learn from Covid-related signals, when appropriate. We are interested in determining whether our model can transfer useful information from the historical model (i.e. \epideep) when required and if it can prevent transfer of detrimental information. 
Specifically, our questions are: 
\par\noindent {\bf Transfer Learning}
\par\noindent Q1. Is \ourmethod{} able to achieve successful positive transfer to model the contamination of wILI values?
\par\noindent Q2. Does~\ourmethod prevent negative transfer by automatically recognizing when wILI and COVID-19 trends deviate? 
\par\noindent {\bf Forecasting Performance}
\par\noindent Q3. Does \ourmethod's emphasis on transfer learning sacrifice overall performance with respect to state-of-the-art methods? 
\par\noindent {\bf Ablation Studies}
\par\noindent Q4. How does each facet of \ourmethod{} affect COVID-ILI forecasting performance?
\par\noindent Q5. What data signals are most relevant to COVID-ILI forecasting?
\par \noindent
Q1 and Q2, which are about transfer learning, are aligned with the main goal of this paper. In Q3, we are interested in determining whether \ourmethod sacrifices any overall forecasting performance, as compared to the state-of-the art (SOTA) baselines, by being too focused on balancing the transfer of knowledge? In Q4 and Q5, we analyze the importance of different components and data signals to performance.

\noindent\textbf{Training and Optimization.} For training \ourmethod, we do the following. We found that, for practical purposes, it is convenient to pre-train \epicn, and then remove its last feedforward layers (decoder). Hence the concatenated output of the RNN encoder and the embedding mapper are input to module \textbf{s}. During joint training, we do not modify \epicn's pre-trained parameters. As recommended in \cite{moon_completely_2017}, we train the HTL architecture in an alternating fashion. 

\par\noindent
\textbf{Baselines.} 
We use traditional historical wILI forecasting methods used in literature~\cite{reich_collaborative_2019}: \epideep, extended \deltadensity from Delphi Group~\cite{brooks_nonmechanistic_2018} (which is SOTA as the top performing method in recent CDC influenza forecasting challenges~\cite{reich_collaborative_2019}), SARIMA from ReichLab~\cite{ray_infectious_2017} (seasonal autoregressive method, top performing in recent CDC challenges),
and \empbayes~\cite{brooks2015flexible} (which leverages transformation of historical seasons to forecast the current season),
and HIST (common persistence baseline which forecasts based on weekly average of the historical seasons).

\subsection{Q1: Leveraging positive transfer for Covid-contaminated wILI}\label{sec:q1_sec}
\label{sec:results_phase1}

\begin{figure}[h]
    \hspace{-0.5cm}
    \begin{tabular}{cc}
    \centering
    \includegraphics[width=0.48\linewidth]{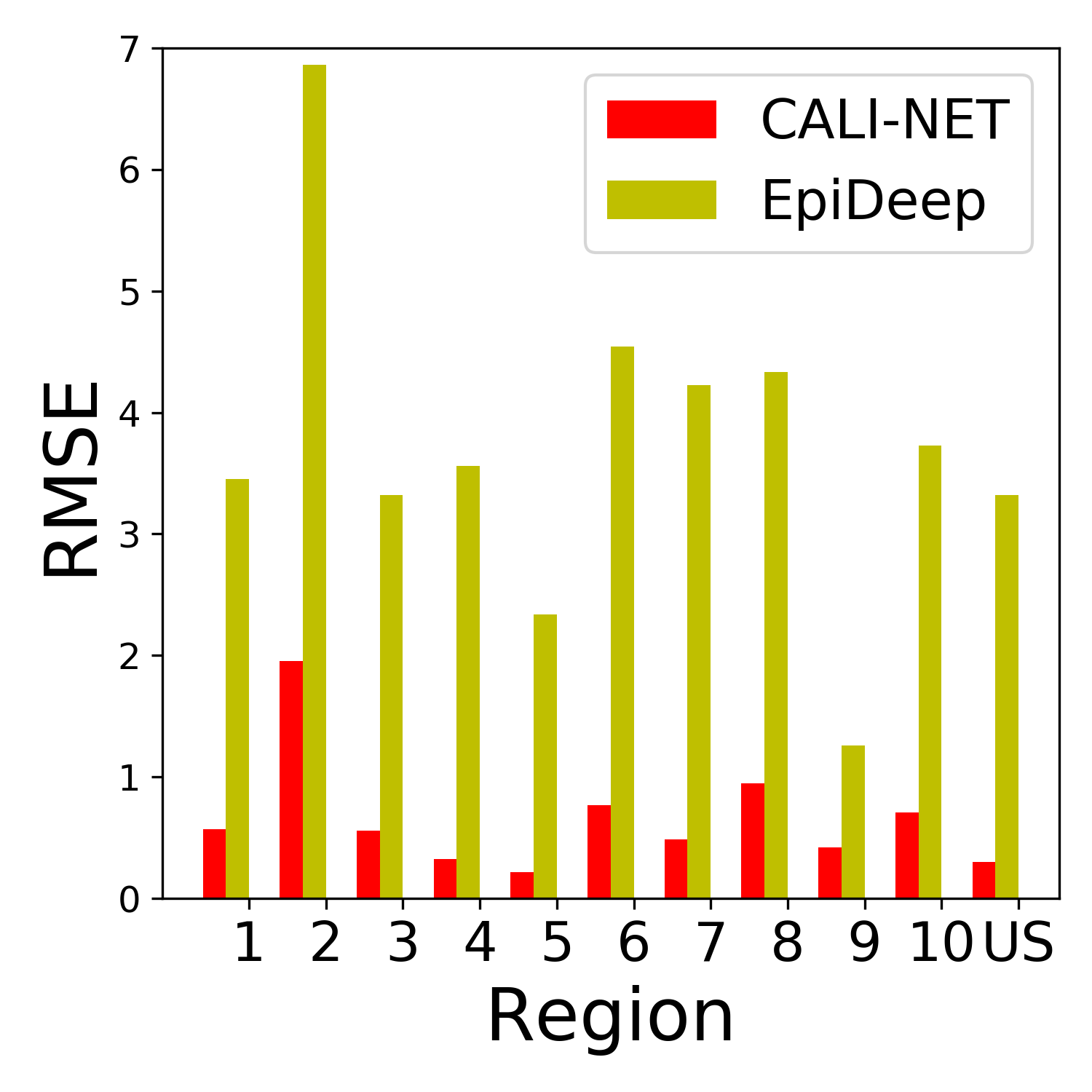} &
     \includegraphics[width=0.48\linewidth]{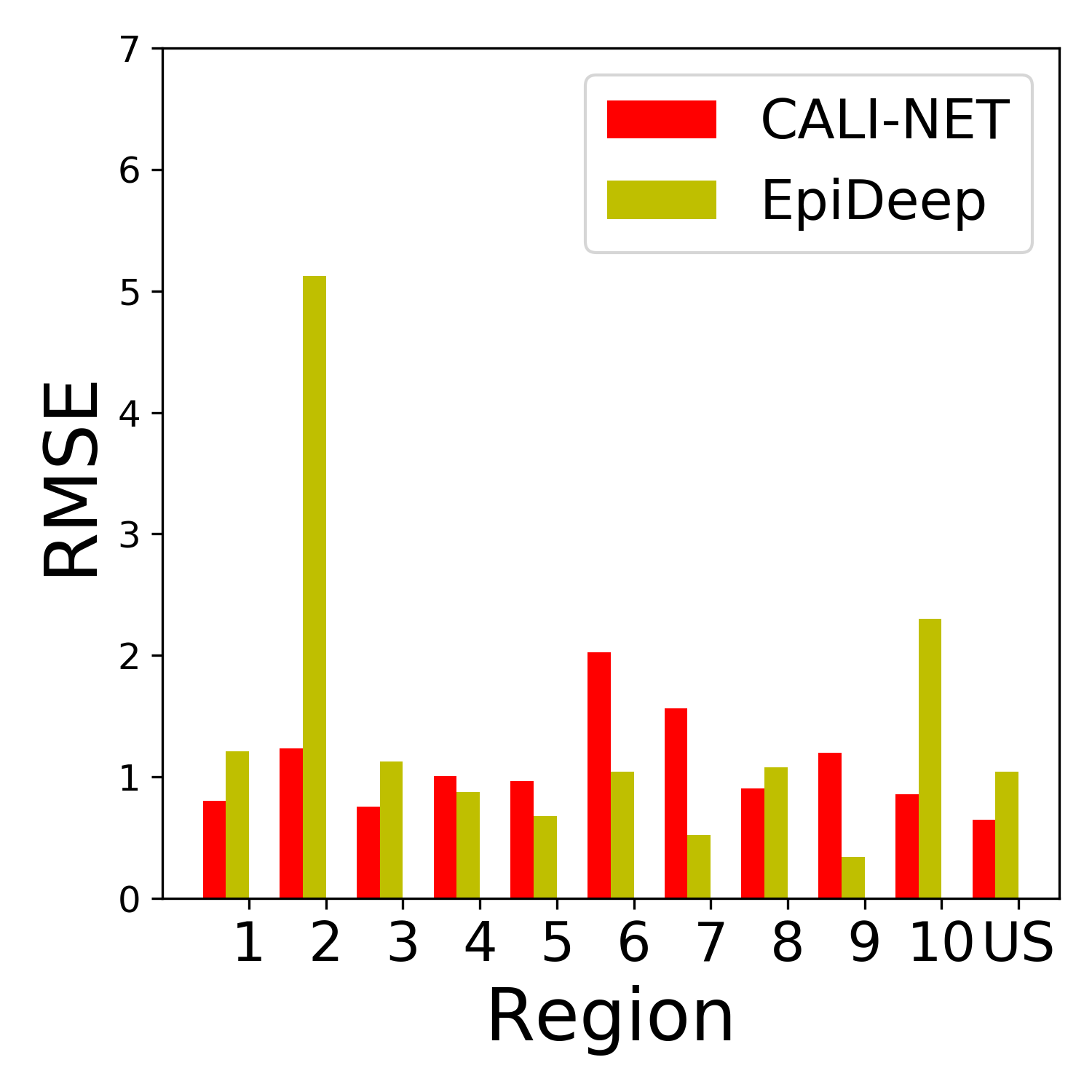} \\
    (a)  Positive transfer stage  & (b) Negative transfer stage \\
  \end{tabular}
    \caption{(a) Our ~\ourmethod framework effectively achieves good forecasts of the uncharacteristic trend in period $T_1$ by steering our influenza forecasting historical model \epideep with knowledge learned from Covid-related signals. (b) Shows forecasting errors from period $T_2$, when the COVID-ILI trend is declining more in tune with the traditional wILI pattern. We notice that \ourmethod is competitive with \epideep, also outperforming it in 6 out of 11 regions while remaining competitive in the rest of the regions. }
    \label{fig:rmse_per_week_phase1}
\end{figure}

The effect of contamination is most pronounced in $T_1$, leading to the COVID-ILI curve exhibiting uncharacteristic non-trivial progression dynamics. 
Therefore, to effectively steer our historical model, \ourmethod automatically leverages positive transfer of Covid-related signals into \epideep. The effect of this automatic positive transfer is shown in Fig.~\ref{fig:rmse_per_week_phase1}(a) where we see that~\ourmethod significantly outperforms \epideep across \emph{all regions}, thanks to our architecture. 

\subsection{Q2: Does~\ourmethod prevent negative transfer automatically?} 
\label{subsec:q2}
Having showcased the adaptation of~\ourmethod in $T_1$, we now show in Fig.~\ref{fig:rmse_per_week_phase1}(b) that our method is effective at preventing negative transfer when wILI is no longer aligned with the exogenous COVID signals (i.e., period $T2$).
In the first place, in some regions the wILI trajectory was never significantly affected by COVID as confirmed COVID cases started to increase significantly only once the influenza season ended.
Second, COVID-affected wILI trajectories of regions displayed a subsequent downtrend after a few weeks. This may be due to the change in care-seeking behavior of outpatients~\cite{kou2020unmasking}.
In this stage, preventing negative transfer from COVID-related signals is needed, such that our model displays more characteristics of traditional influenza models. From Fig.~\ref{fig:rmse_per_week_phase1}(b), we see that \ourmethod is better than \epideep in a majority of the regions indicating that it is able to effectively stop knowledge from misaligned COVID signals from adversely affecting forecasting accuracy thereby effectively preventing negative transfer.

\subsection{Q3: Does \ourmethod sacrifice overall performance?}\label{sec:q3_sec}
\begin{figure}[h]
    \hspace{-0.5cm}
    \centering
    \includegraphics[width=\linewidth]{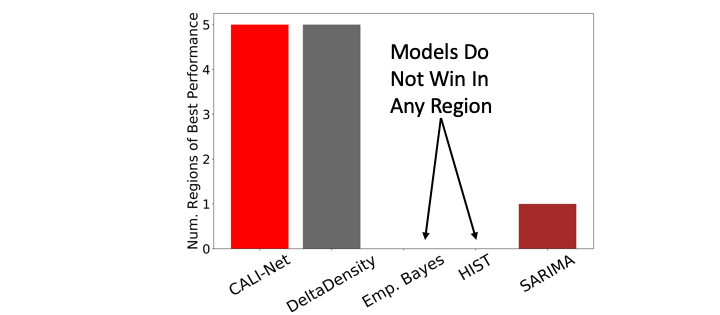}
    \caption{Overall results of \ourmethod compared to Empirical Bayes and SOTA baseline DeltaDensity. We show number of regions in which each model yields best performance and notice \ourmethod outperforms other models in 5 out of 11 regions, on par with DeltaDensity which also yields best performance in 5 other regions with SARIMA being the best in a single region (i.e., Region 9). 
    Models performing within 1\% of the best model per region are considered equivalent best performers.  Hence~\ourmethod yields competitive performance across the entire course $T$. 
    }
    \label{fig:rmse_overall}
\end{figure}

Sec 5.1 and 5.2 show that \ourmethod successfully achieves the main goal of the paper i.e. steering a historical model in a novel scenario. We now study if we sacrifice any performance in this process. To this end, we compare \ourmethod with the traditional SOTA wILI forecasting approaches for the entire course $T$. Specifically, we quantify the number of regions (among all 11), where each method outperforms all others. Fig.~\ref{fig:rmse_overall}  
showcases our findings.
Overall, \ourmethod is able to match the performance of the SOTA historical wILI forecasting models in forecasts for the entire course $T$ and is the top performer in 5 of 11 regions and is one of top 2 best models in 10 out of 11 regions. 
Note that the traditional wILI baselines do not capture the non-seasonal rise of wILI due to COVID contamination in $T_1$  (see Fig.~\ref{fig:intro} and the \supplement). Hence, we note that \ourmethod is the best-suited approach for real-time forecasting in a novel scenario as it captures the non-seasonal patterns while maintaining overall performance. Moreover, we also noticed that \ourmethod outperforms all baselines in regions worst affected by COVID (see \supplement).

\subsection{Q4 and Q5: Module, Data and Parameter Importance and Sensitivity}

\noindent\textbf{Justification for \feature Architecture}.
We conducted an ablation study testing the three components of \feature: (a) regional reconstruction, (b) Laplacian regularization, and (c) the recurrent model.
We found that removing each of them degrades performance showing their individual effectiveness.

\par\noindent
\textbf{Module Performance Analysis.}
The figure below showcases RMSE evolution over period $T_2$ for the
\begin{wrapfigure}{r}{0.28\textwidth}
\vspace{-3pt}
\hskip -1cm
\centering
\includegraphics[width=0.3\textwidth]{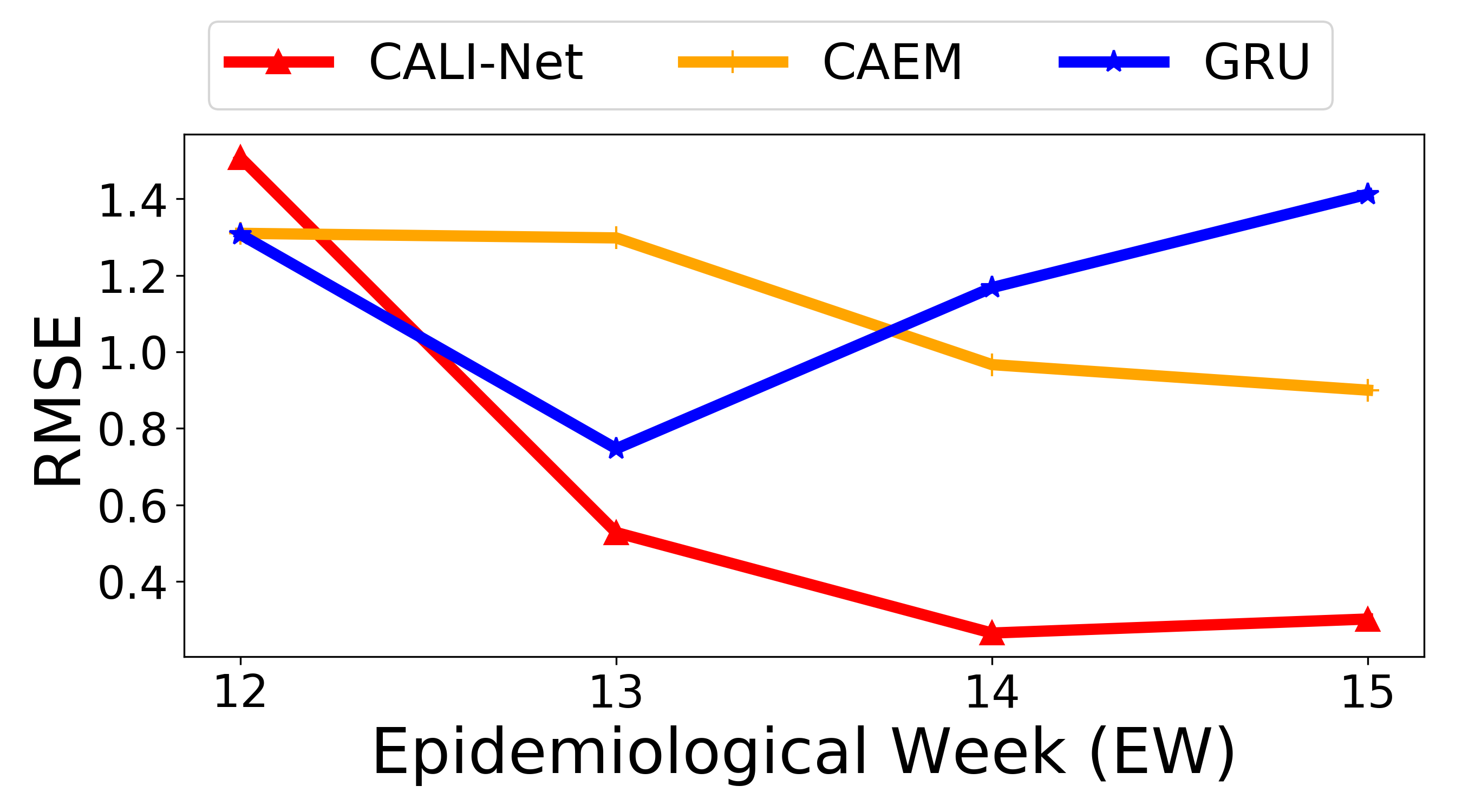}
\label{fig:rmse_per_week}
\end{wrapfigure}
national region for \ourmethod and sub-models of \ourmethod that do not have transfer learning capability. 
Both \gru (standard gated recurrent unit model) and the standalone \feature model use exogenous data as \ourmethod does.
\ourmethod is the only model able to adapt quickly to the downtrend in period $T_2$, due to the effect of the HTL framework which prevents negative transfer of knowledge from COVID related signals, while other models fail to adapt and, in fact, predict rising or flat wILI forecasts.

From Sec.~\ref{sec:q1_sec} - \ref{sec:q3_sec}, we see that \ourmethod is the \emph{only} method capable of capturing both the initial uptrend of COVID-ILI and the subsequent decline effectively, showing its usefulness for emerging diseases. 

\par\noindent\textbf{Effect of KD.} We perform an ablation study to understand the contribution of the proposed KD losses. 
Often the usefulness of source (\epicn) and target (\feature) modules vary depending on the usefulness of the historical and exogenous data sources. Our attentive KD distillation losses provide structure and balance to the transferred knowledge.

We compare 1-week ahead forecasting performance of~\ourmethod{} and a variant of~\ourmethod{} with KD losses removed. 
\begin{wrapfigure}{c}{0.21\textwidth}
\vspace{-1pt}
\hskip -1cm
\centering
    \includegraphics[width=.21\textwidth]{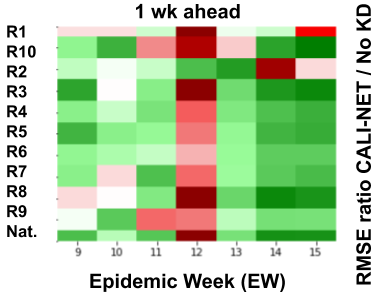}
  \label{fig:per_region_rmse_heatmap}
\end{wrapfigure}
Specifically, see Fig. right; each box is colored by the ratio of RMSE of \ourmethod{} and its variant (capped at -1 and 1 to help visualization). Green cells indicate that~\ourmethod{} does better while red cells indicate that \ourmethod{} w/o KD losses does better. We see that for 1-week ahead forecasting, structuring the knowledge transferred from \epideep proves to be valuable for most EWs. However, for long-term forecasting, KD losses seem to downgrade guidance of \epicn (results in \supplement).
This may be because the season seems to revert to typical behavior in the time-frame predicted in long-term forecasts.

\noindent\textbf{Contribution of Exogenous Signals}. In the figure below, we can see the average overall RMSE obtained when a single data bucket was removed during the training of \ourmethod.
\begin{wrapfigure}{r}{0.17\textwidth}
\hskip -1cm
\centering
\includegraphics[width=0.17\textwidth]{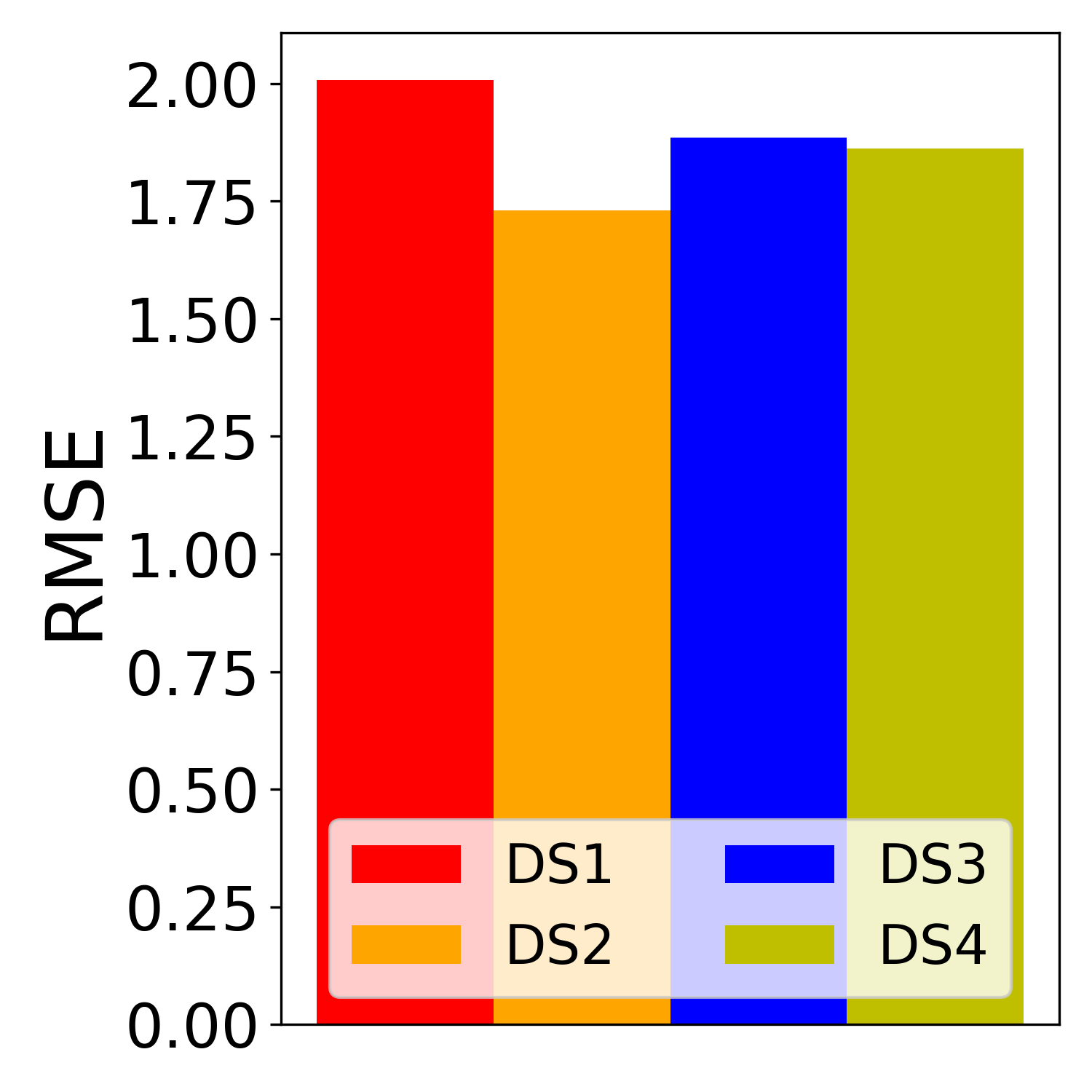}
\label{fig:data_ablation_small}
\vspace{-10pt}
\end{wrapfigure}
We noticed that line list based data (DS1) is very helpful in COVID-ILI forecasting while the effectiveness of testing (DS2) and crowdsourced based (DS3) data is slightly more varied across regions, an observation that resolves Q5. 
This also suggests that data closer to the disease is more reliable. 
More detailed results (regional breakdown) are in the \supplement. 

\noindent\textbf{Parameter Sensitivity}.
For the hyperparameters of \ourmethod, we perform thorough experiments and demonstrate the robustness of our method.
Details in the \supplement.

\section{Discussion}
Here we introduced the challenging COVID-ILI forecasting task, and proposed our novel approach \ourmethod.  We show the usefulness of a principled method to transfer relevant knowledge from an existing deep flu forecasting model (based on rich historical data) to one relying on relevant but limited recent COVID-related exogenous signals. Our method is based on carefully designed components to avoid negative transfer (by attentive KD losses), promote spatial consistency (via Laplacian losses in a novel recurrent architecture \feature), and also handle data paucity (via the global nature of \feature and other aspects).
\ourmethod effectively captures non-trivial atypical trends in COVID-ILI  evolution whereas other models and baselines do not. We also demonstrate how each of our components and data signals is important and useful for performance. These results provide guidance for steering forecasting models in an emerging disease scenario.
 In future, we believe our techniques can be applied to other source models (in addition to \epicn), as well as designing more sophisticated architectures for the target \feature model. We can also  explore adding interpretability to our forecasts for additional insights.

\section*{Acknowledgments}
This paper is based on work partially supported by the NSF (Expeditions CCF-1918770, CAREER IIS-2028586, RAPID IIS-2027862, Medium IIS-1955883, NRT DGE-1545362, IIS-1633363, OAC-1835660), CDC MInD program, ORNL and funds/computing resources from Georgia Tech and GTRI.
B. A. was in part supported by the CDC MInD-Healthcare U01CK000531-Supplement.


\small

\begin{thebibliography}{46}
\providecommand{\natexlab}[1]{#1}
\providecommand{\url}[1]{\texttt{#1}}
\providecommand{\urlprefix}{URL }
\expandafter\ifx\csname urlstyle\endcsname\relax
  \providecommand{\doi}[1]{doi:\discretionary{}{}{}#1}\else
  \providecommand{\doi}{doi:\discretionary{}{}{}\begingroup
  \urlstyle{rm}\Url}\fi

\bibitem[{Adhikari et~al.(2019)Adhikari, Xu, Ramakrishnan, and
  Prakash}]{adhikari2019epideep}
Adhikari, B.; Xu, X.; Ramakrishnan, N.; and Prakash, B.~A. 2019.
\newblock Epideep: Exploiting embeddings for epidemic forecasting.
\newblock In \emph{Proceedings of the 25th ACM SIGKDD International Conference
  on Knowledge Discovery \& Data Mining}, 577--586.

\bibitem[{Ba and Caruana(2014)}]{ba2014deep}
Ba, J.; and Caruana, R. 2014.
\newblock Do deep nets really need to be deep?
\newblock In \emph{Advances in neural information processing systems},
  2654--2662.

\bibitem[{Belkin, Matveeva, and Niyogi(2004)}]{belkin2004regularization}
Belkin, M.; Matveeva, I.; and Niyogi, P. 2004.
\newblock Regularization and semi-supervised learning on large graphs.
\newblock In \emph{International Conference on Computational Learning Theory},
  624--638. Springer.

\bibitem[{Biggerstaff et~al.(2016)Biggerstaff, Alper, Dredze, Fox, Fung,
  Hickmann, Lewis, Rosenfeld, Shaman, Tsou et~al.}]{biggerstaff2016results}
Biggerstaff, M.; Alper, D.; Dredze, M.; Fox, S.; Fung, I. C.-H.; Hickmann,
  K.~S.; Lewis, B.; Rosenfeld, R.; Shaman, J.; Tsou, M.-H.; et~al. 2016.
\newblock Results from the centers for disease control and prevention’s
  predict the 2013--2014 Influenza Season Challenge.
\newblock \emph{BMC infectious diseases} 16(1): 357.

\bibitem[{Bo{\"e}lle et~al.(2020)Bo{\"e}lle, Souty, Launay, Guerrisi, Turbelin,
  Behillil, Enouf, Poletto, Lina, van~der Werf et~al.}]{boelle2020excess}
Bo{\"e}lle, P.-Y.; Souty, C.; Launay, T.; Guerrisi, C.; Turbelin, C.; Behillil,
  S.; Enouf, V.; Poletto, C.; Lina, B.; van~der Werf, S.; et~al. 2020.
\newblock Excess cases of influenza-like illnesses synchronous with coronavirus
  disease (COVID-19) epidemic, France, March 2020.
\newblock \emph{Eurosurveillance} 25(14): 2000326.

\bibitem[{Box et~al.(2015)Box, Jenkins, Reinsel, and Ljung}]{box2015time}
Box, G.~E.; Jenkins, G.~M.; Reinsel, G.~C.; and Ljung, G.~M. 2015.
\newblock \emph{Time series analysis: forecasting and control}.
\newblock John Wiley \& Sons.

\bibitem[{Brooks et~al.(2015)Brooks, Farrow, Hyun, Tibshirani, and
  Rosenfeld}]{brooks2015flexible}
Brooks, L.~C.; Farrow, D.~C.; Hyun, S.; Tibshirani, R.~J.; and Rosenfeld, R.
  2015.
\newblock Flexible modeling of epidemics with an empirical Bayes framework.
\newblock \emph{PLoS computational biology} 11(8): e1004382.

\bibitem[{Brooks et~al.(2018)Brooks, Farrow, Hyun, Tibshirani, and
  Rosenfeld}]{brooks_nonmechanistic_2018}
Brooks, L.~C.; Farrow, D.~C.; Hyun, S.; Tibshirani, R.~J.; and Rosenfeld, R.
  2018.
\newblock Nonmechanistic forecasts of seasonal influenza with iterative
  one-week-ahead distributions.
\newblock \emph{PLOS Computational Biology} 14(6): e1006134.
\newblock ISSN 1553-7358.
\newblock \doi{10.1371/journal.pcbi.1006134}.

\bibitem[{Castrofino et~al.(2020)Castrofino, Del~Castillo, Grosso, Barone,
  Gramegna, Galli, Tirani, Castaldi, Pariani, and
  Cereda}]{castrofino2020influenza}
Castrofino, A.; Del~Castillo, G.; Grosso, F.; Barone, A.; Gramegna, M.; Galli,
  C.; Tirani, M.; Castaldi, S.; Pariani, E.; and Cereda, D. 2020.
\newblock Influenza surveillance system and Covid-19.
\newblock \emph{European Journal of Public Health} 30(Supplement\_5):
  ckaa165--354.

\bibitem[{CDC(2020)}]{cdc2020}
CDC. 2020.
\newblock Weekly U.S. Influenza Surveillance Report.
\newblock \urlprefix\url{https://cdc.gov/flu/weekly/index.html}.

\bibitem[{Chakraborty et~al.(2014)Chakraborty, Khadivi, Lewis, Mahendiran,
  Chen, Butler, Nsoesie, Mekaru, Brownstein, Marathe
  et~al.}]{chakraborty2014forecasting}
Chakraborty, P.; Khadivi, P.; Lewis, B.; Mahendiran, A.; Chen, J.; Butler, P.;
  Nsoesie, E.~O.; Mekaru, S.~R.; Brownstein, J.~S.; Marathe, M.~V.; et~al.
  2014.
\newblock Forecasting a moving target: Ensemble models for ILI case count
  predictions.
\newblock In \emph{Proceedings of the 2014 SIAM international conference on
  data mining}, 262--270. SIAM.

\bibitem[{Chen et~al.(2016)Chen, Hossain, Butler, Ramakrishnan, and
  Prakash}]{chen2016syndromic}
Chen, L.; Hossain, K.~T.; Butler, P.; Ramakrishnan, N.; and Prakash, B.~A.
  2016.
\newblock Syndromic surveillance of Flu on Twitter using weakly supervised
  temporal topic models.
\newblock \emph{Data mining and knowledge discovery} 30(3): 681--710.

\bibitem[{Cho et~al.(2014)Cho, Van~Merri{\"e}nboer, Gulcehre, Bahdanau,
  Bougares, Schwenk, and Bengio}]{cho2014learning}
Cho, K.; Van~Merri{\"e}nboer, B.; Gulcehre, C.; Bahdanau, D.; Bougares, F.;
  Schwenk, H.; and Bengio, Y. 2014.
\newblock Learning phrase representations using RNN encoder-decoder for
  statistical machine translation.
\newblock \emph{arXiv preprint arXiv:1406.1078} .

\bibitem[{Connor, Martin, and Atlas(1994)}]{connor1994recurrent}
Connor, J.~T.; Martin, R.~D.; and Atlas, L.~E. 1994.
\newblock Recurrent neural networks and robust time series prediction.
\newblock \emph{IEEE transactions on neural networks} 5(2): 240--254.

\bibitem[{COVID-Tracking(2020)}]{covidtracking}
COVID-Tracking. 2020.
\newblock The COVID Tracking Project.
\newblock \urlprefix\url{https://covidtracking.com}.

\bibitem[{Dredze et~al.(2014)Dredze, Cheng, Paul, and
  Broniatowski}]{dredze2014healthtweets}
Dredze, M.; Cheng, R.; Paul, M.~J.; and Broniatowski, D. 2014.
\newblock HealthTweets. org: a platform for public health surveillance using
  Twitter.
\newblock In \emph{Workshops at the Twenty-Eighth AAAI Conference on Artificial
  Intelligence}.

\bibitem[{Fu, Zhang, and Li(2016)}]{fu2016using}
Fu, R.; Zhang, Z.; and Li, L. 2016.
\newblock Using LSTM and GRU neural network methods for traffic flow
  prediction.
\newblock In \emph{2016 31st Youth Academic Annual Conference of Chinese
  Association of Automation (YAC)}, 324--328. IEEE.

\bibitem[{Ginsberg et~al.(2009)Ginsberg, Mohebbi, Patel, Brammer, Smolinski,
  and Brilliant}]{ginsberg2009detecting}
Ginsberg, J.; Mohebbi, M.~H.; Patel, R.~S.; Brammer, L.; Smolinski, M.~S.; and
  Brilliant, L. 2009.
\newblock Detecting influenza epidemics using search engine query data.
\newblock \emph{Nature} 457(7232): 1012.

\bibitem[{Hinton, Vinyals, and Dean(2015)}]{hinton2015distilling}
Hinton, G.; Vinyals, O.; and Dean, J. 2015.
\newblock Distilling the knowledge in a neural network.
\newblock \emph{arXiv preprint arXiv:1503.02531} .

\bibitem[{Hochreiter and Schmidhuber(1997)}]{hochreiter1997long}
Hochreiter, S.; and Schmidhuber, J. 1997.
\newblock Long short-term memory.
\newblock \emph{Neural computation} 9(8): 1735--1780.

\bibitem[{JHU(2020)}]{jhu-covid}
JHU. 2020.
\newblock JHU CSSE COVID-19 Dashboard.
\newblock \urlprefix\url{https://coronavirus.jhu.edu/map.html}.

\bibitem[{Kou et~al.(2020)Kou, Yang, Chang, Ho, and Graver}]{kou2020unmasking}
Kou, S.; Yang, S.; Chang, C.-J.; Ho, T.-H.; and Graver, L. 2020.
\newblock Unmasking the Actual COVID-19 Case Count.
\newblock \emph{Clinical Infectious Diseases} .

\bibitem[{Lee, Agrawal, and Choudhary(2013)}]{lee2013real}
Lee, K.; Agrawal, A.; and Choudhary, A. 2013.
\newblock Real-time disease surveillance using twitter data: demonstration on
  flu and cancer.
\newblock In \emph{Proceedings of the 19th ACM SIGKDD international conference
  on Knowledge discovery and data mining}, 1474--1477. ACM.

\bibitem[{Li et~al.(2019)Li, Pan, Wan, and Kot}]{li_heterogeneous_2019}
Li, H.; Pan, S.~J.; Wan, R.; and Kot, A.~C. 2019.
\newblock Heterogeneous {Transfer} {Learning} via {Deep} {Matrix} {Completion}
  with {Adversarial} {Kernel} {Embedding}.
\newblock \emph{Proceedings of the AAAI Conference on Artificial Intelligence}
  33: 8602--8609.
\newblock ISSN 2374-3468, 2159-5399.
\newblock \doi{10.1609/aaai.v33i01.33018602}.

\bibitem[{Miller et~al.(2018)Miller, Singh, Koehler, and
  Polgreen}]{miller2018smartphone}
Miller, A.~C.; Singh, I.; Koehler, E.; and Polgreen, P.~M. 2018.
\newblock A smartphone-driven thermometer application for real-time
  population-and individual-level influenza surveillance.
\newblock \emph{Clinical Infectious Diseases} 67(3): 388--397.

\bibitem[{Moon and Carbonell(2017)}]{moon_completely_2017}
Moon, S.; and Carbonell, J.~G. 2017.
\newblock Completely {Heterogeneous} {Transfer} {Learning} with
  {Attention}-{What} {And} {What} {Not} {To} {Transfer}.
\newblock In \emph{Proceedings of the 26th International Joint Conference on
  Artificial Intelligence}, volume~1, 2508--2514. AAAI Press.

\bibitem[{Muralidhar, Muthiah, and Ramakrishnan(2019)}]{muralidhar2019dyat}
Muralidhar, N.; Muthiah, S.; and Ramakrishnan, N. 2019.
\newblock DyAt nets: dynamic attention networks for state forecasting in
  cyber-physical systems.
\newblock In \emph{Proceedings of the 28th International Joint Conference on
  Artificial Intelligence}, 3180--3186. AAAI Press.

\bibitem[{Osthus et~al.(2019)Osthus, Gattiker, Priedhorsky, Del~Valle
  et~al.}]{osthus2019dynamic}
Osthus, D.; Gattiker, J.; Priedhorsky, R.; Del~Valle, S.~Y.; et~al. 2019.
\newblock Dynamic Bayesian influenza forecasting in the United States with
  hierarchical discrepancy (with discussion).
\newblock \emph{Bayesian Analysis} 14(1): 261--312.

\bibitem[{Ray et~al.(2017)Ray, Sakrejda, Lauer, Johansson, and
  Reich}]{ray_infectious_2017}
Ray, E.~L.; Sakrejda, K.; Lauer, S.~A.; Johansson, M.~A.; and Reich, N.~G.
  2017.
\newblock Infectious disease prediction with kernel conditional density
  estimation: {Infectious} disease prediction with kernel conditional density
  estimation.
\newblock \emph{Statistics in Medicine} 36(30): 4908--4929.
\newblock ISSN 02776715.
\newblock \doi{10.1002/sim.7488}.

\bibitem[{Reich et~al.(2019{\natexlab{a}})Reich, Brooks, Fox, Kandula, McGowan,
  Moore, Osthus, Ray, Tushar, Yamana, Biggerstaff, Johansson, Rosenfeld, and
  Shaman}]{reich_collaborative_2019}
Reich, N.~G.; Brooks, L.~C.; Fox, S.~J.; Kandula, S.; McGowan, C.~J.; Moore,
  E.; Osthus, D.; Ray, E.~L.; Tushar, A.; Yamana, T.~K.; Biggerstaff, M.;
  Johansson, M.~A.; Rosenfeld, R.; and Shaman, J. 2019{\natexlab{a}}.
\newblock A collaborative multiyear, multimodel assessment of seasonal
  influenza forecasting in the {United} {States}.
\newblock \emph{Proceedings of the National Academy of Sciences} 201812594.
\newblock ISSN 0027-8424, 1091-6490.
\newblock \doi{10.1073/pnas.1812594116}.

\bibitem[{Reich et~al.(2019{\natexlab{b}})Reich, McGowan, Yamana, Tushar, Ray,
  Osthus, Kandula, Brooks, Crawford-Crudell, Gibson et~al.}]{reich2019accuracy}
Reich, N.~G.; McGowan, C.~J.; Yamana, T.~K.; Tushar, A.; Ray, E.~L.; Osthus,
  D.; Kandula, S.; Brooks, L.~C.; Crawford-Crudell, W.; Gibson, G.~C.; et~al.
  2019{\natexlab{b}}.
\newblock Accuracy of real-time multi-model ensemble forecasts for seasonal
  influenza in the US.
\newblock \emph{PLoS computational biology} 15(11).

\bibitem[{Rodriguez et~al.(2020)Rodriguez, Tabassum, Cui, Xie, Ho, Agarwal,
  Adhikari, and Prakash}]{rodriguez2020deepcovid}
Rodriguez, A.; Tabassum, A.; Cui, J.; Xie, J.; Ho, J.; Agarwal, P.; Adhikari,
  B.; and Prakash, B.~A. 2020.
\newblock DeepCOVID: An Operational Deep Learning-driven Framework for
  Explainable Real-time COVID-19 Forecasting.
\newblock \emph{medRxiv} .

\bibitem[{Sapankevych and Sankar(2009)}]{sapankevych2009time}
Sapankevych, N.~I.; and Sankar, R. 2009.
\newblock Time series prediction using support vector machines: a survey.
\newblock \emph{IEEE Computational Intelligence Magazine} 4(2).

\bibitem[{Saputra et~al.(2019)Saputra, de~Gusmao, Almalioglu, Markham, and
  Trigoni}]{saputra2019distilling}
Saputra, M. R.~U.; de~Gusmao, P.~P.; Almalioglu, Y.; Markham, A.; and Trigoni,
  N. 2019.
\newblock Distilling knowledge from a deep pose regressor network.
\newblock In \emph{Proceedings of the IEEE International Conference on Computer
  Vision}, 263--272.

\bibitem[{Shaman, Goldstein, and Lipsitch(2010)}]{shaman2010absolute}
Shaman, J.; Goldstein, E.; and Lipsitch, M. 2010.
\newblock Absolute humidity and pandemic versus epidemic influenza.
\newblock \emph{American journal of epidemiology} 173(2): 127--135.

\bibitem[{Shaman and Karspeck(2012)}]{shaman2012forecasting}
Shaman, J.; and Karspeck, A. 2012.
\newblock Forecasting seasonal outbreaks of influenza.
\newblock \emph{Proceedings of the National Academy of Sciences} 109(50):
  20425--20430.

\bibitem[{Subbian and Banerjee(2013)}]{subbian2013climate}
Subbian, K.; and Banerjee, A. 2013.
\newblock Climate multi-model regression using spatial smoothing.
\newblock In \emph{Proceedings of the 2013 SIAM International Conference on
  Data Mining}, 324--332. SIAM.

\bibitem[{Takamoto, Morishita, and Imaoka(2020)}]{takamoto2020efficient}
Takamoto, M.; Morishita, Y.; and Imaoka, H. 2020.
\newblock An Efficient Method of Training Small Models for Regression Problems
  with Knowledge Distillation.
\newblock \emph{arXiv preprint arXiv:2002.12597} .

\bibitem[{Tamerius et~al.(2013)Tamerius, Shaman, Alonso, Bloom-Feshbach, Uejio,
  Comrie, and Viboud}]{tamerius2013environmental}
Tamerius, J.~D.; Shaman, J.; Alonso, W.~J.; Bloom-Feshbach, K.; Uejio, C.~K.;
  Comrie, A.; and Viboud, C. 2013.
\newblock Environmental predictors of seasonal influenza epidemics across
  temperate and tropical climates.
\newblock \emph{PLoS pathogens} 9(3): e1003194.

\bibitem[{Tizzoni et~al.(2012)Tizzoni, Bajardi, Poletto, Ramasco, Balcan,
  Gon{\c{c}}alves, Perra, Colizza, and Vespignani}]{tizzoni2012real}
Tizzoni, M.; Bajardi, P.; Poletto, C.; Ramasco, J.~J.; Balcan, D.;
  Gon{\c{c}}alves, B.; Perra, N.; Colizza, V.; and Vespignani, A. 2012.
\newblock Real-time numerical forecast of global epidemic spreading: case study
  of 2009 A/H1N1pdm.
\newblock \emph{BMC medicine} 10(1): 165.

\bibitem[{Venna et~al.(2017)Venna, Tavanaei, Gottumukkala, Raghavan, Maida, and
  Nichols}]{venna2017novel}
Venna, S.~R.; Tavanaei, A.; Gottumukkala, R.~N.; Raghavan, V.~V.; Maida, A.;
  and Nichols, S. 2017.
\newblock A novel data-driven model for real-time influenza forecasting.
\newblock \emph{bioRxiv} 185512.

\bibitem[{Volkova et~al.(2017)Volkova, Ayton, Porterfield, and
  Corley}]{volkova2017forecasting}
Volkova, S.; Ayton, E.; Porterfield, K.; and Corley, C.~D. 2017.
\newblock Forecasting influenza-like illness dynamics for military populations
  using neural networks and social media.
\newblock \emph{PloS one} 12(12): e0188941.

\bibitem[{Wang, Chen, and Marathe(2019)}]{wang2019defsi}
Wang, L.; Chen, J.; and Marathe, M. 2019.
\newblock DEFSI: Deep learning based epidemic forecasting with synthetic
  information.
\newblock In \emph{Proceedings of the AAAI Conference on Artificial
  Intelligence}, volume~33, 9607--9612.

\bibitem[{Yan et~al.(2018)Yan, Li, Wu, Min, Tan, and
  Wu}]{yan_semi-supervised_2018}
Yan, Y.; Li, W.; Wu, H.; Min, H.; Tan, M.; and Wu, Q. 2018.
\newblock Semi-{Supervised} {Optimal} {Transport} for {Heterogeneous} {Domain}
  {Adaptation}.
\newblock In \emph{Proceedings of the {Twenty}-{Seventh} {International}
  {Joint} {Conference} on {Artificial} {Intelligence}}, 2969--2975. Stockholm,
  Sweden: International Joint Conferences on Artificial Intelligence
  Organization.
\newblock ISBN 978-0-9992411-2-7.
\newblock \doi{10.24963/ijcai.2018/412}.

\bibitem[{Yuan et~al.(2013)Yuan, Nsoesie, Lv, Peng, Chunara, and
  Brownstein}]{yuan2013monitoring}
Yuan, Q.; Nsoesie, E.~O.; Lv, B.; Peng, G.; Chunara, R.; and Brownstein, J.~S.
  2013.
\newblock Monitoring influenza epidemics in china with search query from baidu.
\newblock \emph{PloS one} 8(5): e64323.

\bibitem[{Zhang et~al.(2017)Zhang, Perra, Perrotta, Tizzoni, Paolotti, and
  Vespignani}]{zhang2017forecasting}
Zhang, Q.; Perra, N.; Perrotta, D.; Tizzoni, M.; Paolotti, D.; and Vespignani,
  A. 2017.
\newblock Forecasting seasonal influenza fusing digital indicators and a
  mechanistic disease model.
\newblock In \emph{Proceedings of the 26th International Conference on World
  Wide Web}, 311--319. International World Wide Web Conferences Steering
  Committee.

\end{thebibliography}
\end{document}